\def\year{2019}\relax
\documentclass[letterpaper]{article} 
\usepackage{aaai19}  
\usepackage{times}  
\usepackage{helvet}  
\usepackage{courier}  
\usepackage{url}  
\usepackage{graphicx}  
\usepackage{times}
\usepackage{latexsym}
\usepackage{times}
\usepackage{url}
\usepackage{latexsym}
\usepackage{graphicx}
\usepackage{multirow}
\usepackage{amsmath,amsfonts,amsthm,bm}
\usepackage{graphicx}
\usepackage{capt-of}
\usepackage{multirow}
\usepackage{listings, todonotes}
\usepackage{tabularx}
\usepackage{color}
    \definecolor{dkgreen}{rgb}{0,0.6,0}
    \definecolor{gray}{rgb}{0.5,0.5,0.5}
    \definecolor{mauve}{rgb}{0.58,0,0.82}
\usepackage{url}
\newcommand{\SO}[0]{Stack Overflow}
\newcommand{\Comment}[1]{}
\newcommand{\Space}[1]{}
\newcommand{\Ku}[0]{Knowledge unit}
\newcommand{\ku}[0]{knowledge unit}
\newcommand{\kua}[0]{KU}

\newcommand{\Fix}[1]{}
\newcommand{\FFix}[1]{}

\newcommand{\Our}[0]{Our}

\newcommand{\softgoog}[0]{Soft\_Google}
\newcommand{\softSO}[0]{Soft\_SO}
\newcommand{\softLev}[0]{Soft\_Lev}
\newcommand{\SSVM}[0]{\textsc{SoftSVM}}
\newcommand{\dotbilstm}[0]{\textsc{DotBiLSTM}}

\newcommand{\Part}[1]{\noindent\emph{#1 }}

\date{}
\begin{document}
%
\title{Question Relatedness on \SO{}: The Task, Dataset, and Corpus-inspired Models}


\author{Amirreza Shirani$^{\dagger}$, 
Bowen Xu$^{\ddagger}$, 
David Lo$^{\ddagger}$,
Thamar Solorio$^{\dagger}$ and 
Amin Alipour$^{\dagger}$ \\
 	$^{\dagger}$University of Houston\\
  	$^{\ddagger}$Singapore Management University
}

\maketitle
\begin{abstract}

Domain-specific community question answering is becoming an integral part of professions.
Finding related questions and answers in these communities can significantly improve the effectiveness and efficiency of information seeking. 
\SO{} is one of the most popular communities that is being used by millions of programmers. 
In this paper, we analyze the problem of predicting \ku{} (question thread) relatedness in \SO{}. 
In particular, 
we formulate the question relatedness task as a multi-class classification problem with
four degrees of relatedness.

We present a large-scale dataset with more than $300K$ pairs.
To the best of our knowledge, this dataset is the largest domain-specific dataset for Question-Question relatedness. 
We present the steps that we took to collect, clean, process, and assure the quality of the dataset.
The proposed dataset on \SO{} is a useful resource to develop novel solutions, specifically data-hungry neural network models, for the prediction of relatedness in technical community question-answering forums.

We adapt a neural network architecture and a traditional model for this task that effectively utilize information from different parts of \ku{}s to compute the relatedness between them.
These models can be used to benchmark novel models, as they perform well in our task and in a closely similar task.
\end{abstract}

\section{Introduction}
\Comment{
    \Fix{Amin: not an interesting opening}
    Community question answering (cQA) is a popular research topic in Natural Language Processing (NLP).
    Many NLP problems related to cQA have been studied, for example, answer summarization~\cite{chan2012community,demner2006answer,liu2008understanding}, question answer matching~\cite{tan2016improved,shen2015question,ferret2001terminological} and question semantic matching~\cite{bogdanova2015detecting,wu2011efficient,nakov2017semeval}.
    In addition, some datasets are built for NLP tasks to evaluate NLP approaches, for example, TREC QA Dataset for QA Retrieval and Ranking~\cite{wang2007jeopardy}, Yahoo! Answers Dataset for  Question Answer retrieval~\cite{qiu2015convolutional}, Quora Dataset for Question Semantic Matching~\cite{quoradataset}.
}

Community question answering (cQA) is becoming an integral part of professions allowing users to tap on crowds' wisdom and find answers to their questions.
Techniques, such as answer summarization~\cite{chan2012community,xu2017answerbot,demner2006answer,liu2008understanding}, question answer matching~\cite{tan2016improved,shen2015question} and question semantic matching~\cite{bogdanova2015detecting,wu2011efficient,nakov2017semeval}, have been devised to improve users' experience by accelerating  finding relevant information and enhancing the information presentation to users.

We refer to the collection of a question along with all its answers as a \emph{\ku{}} (KU).
Finding related \ku{} in these communities can significantly improve the effectiveness and efficiency of information seeking.
It allows users to navigate between \ku{}s, prune unrelated \ku{}s from the information search space.
Finding related \ku{}s can be quite time-consuming due to the fact that even the same question can be rephrased in many different ways. 
Therefore automated techniques to identify related \ku{}s are desirable.

In this work, we describe the task of prediction of \emph{relatedness}  in \SO{}\Space{\footnote{Stack Overflow, \url{https://stackoverflow.com/}}}, the most popular resource for topics related to software development.
Knowledge in \SO{} is dispersed and developers usually need to explore several related \ku{}s to gain insights into the problem at hand and possible solutions.
\SO{} has become an indispensable tool for programmers; about 50 million developers visit it monthly, and over 85\% of users visit \SO{} almost daily.\footnote{Stack Overflow 2018 Developer Survey, \url{https://insights.stackoverflow.com/survey/2018/}}
The reputation of this webiste has attracted many developers to actively participate and contribute to the forum. A study showed that most questions on \SO{} are answered within 11 minutes of posting them~\cite{Mamykina:2011:CHI}. 

We formulate the problem of identification of related \kua{}s, as a multi-class classification problem by breaking relatedness into multiple classes.
More precisely, a model has to classify the degree of relatedness of two KUs into one of four classes: \emph{duplicate}, \emph{direct}, \emph{indirect}, or \emph{isolated}.

Predicting relatedness in \SO{} poses an interesting challenge because in addition to natural text, \kua{}s contain a huge amount of programming terms which is of a different nature, and  like many other cQA websites, different users exhibit different discursive habits in posting questions and answers; e.g., some provide minimal details in their questions or answers, while some tend to include a sizable amount of information.

We create a large, reliable dataset for training and testing models for this task.
It contains more than 300K \ku{} pairs annotated with their corresponding relatedness class.
We report all steps to collect, clean, process, and assure the quality of the dataset. 
We rely on URL sharing in \SO{} to decide on the relatedness of KUs, as that programmers facing a specific problem are the best ones to judge the degree of relatedness of questions.
We verified the reliability of our approach by conducting a user study.

\Comment{
\Fix{REza: I added the following because Thamar mentioned that it is not clear how we annotate the dataset}
\Fix{
    \SO{} users discuss a variety of topics, therefore, data annotation process requires a large number of annotators, expert in different areas\Fix{Reza: rephrase!}. 
    We believe that programmers facing a specific problem are the best ones to judge the degree of relatedness of questions.
    In this task, we rely on URL sharing throughout the whole network as it provides information about the relatedness of questions. }
}

To establish a baseline for future evaluations, we present two successful neural network and traditional machine learning models.
we adapt a lightweight Bidirectional Long Short-term Memory (BiLSTM) model tailored to our proposed dataset. 
We also investigate so-called soft-cosine similarity features in a Support Vector Machine (SVM) model.
To investigate the adequacy of these models, we evaluate them on a closely related duplicate detection task.
Our experiments show that our models outperform the state-of-the-art techniques in a duplicate detection task, suggesting that our models are potent benchmarks for our task. 
\Comment{
    Amin: this doesn't read well.
    Interestingly, by systematizing four fine-grained relatedness classes (duplicate, directed, undirected, and isolated) more information is afforded about existing relations between \ku{}s than binary duplicate detection.
    Classes can also merge to constitute larger classes, e.g. related and unrelated. 
}

\noindent \textbf{Contributions.} This paper makes the following contributions.
\begin{itemize}
    \item We present the task of question relatedness in \SO{}, with four degrees of similarity. 
    \item We present a reliable, large dataset for \ku{}s relatedness in \SO{}.
    \item We adapt a corpus-inspired BiLSTM architecture for relatedness detection.
    \item We evaluate the performance of SVM models with several hand-crafted features to predict the relatedness in \SO{}.

\end{itemize}

\section{Related Work}
There are several tasks related to identifying semantically
relevant questions such as
Duplicate Question Detection (DQD), Question-Question similarity, and paraphrase identification.

Perhaps, one of the best-known general-domain DQD dataset is Quora~\footnote{https://goo.gl/kWCcD4} with more than 400K question pairs. Quora dataset was released on Kaggle competition platform in January 2017.
Most of the questions on Quora are asked in one piece without any further description and are not restricted to any domain.
Another well-known DQD dataset is AskUbuntu~\cite{rodrigues2017ways}.
Similar to our \SO{} dataset, AskUbuntu dataset is acquired from Stack Exchange data dump \footnote{https://askubuntu.com/} (September 2014).
The differences are that AskUbuntu dataset only provides binary classes (DQD), it is 11 times smaller than our proposed dataset and only consist of titles and bodies in a concatenated form. 
Many solutions are proposed to address the DQD problem.
\cite{bogdanova2015detecting} utilized a convolutional neural network (CNN) to address the DQD problem on AskUbuntu and Meta datasets.
\cite{silva2020} applied the same model on the cleaned version of datasets and showed that after removing Stack Exchange clues, the results drop by 20\%. 
A more advanced architecture introduced in \cite{rodrigues2017ways} on AskUbuntu and Quora datasets.
This model can be considered as the state-of-the-art model on AskUbuntu dataset which utilizes the combination of a \emph{MayoNLP} model introduced in~\cite{afzal2016mayonlp} and a \emph{CNN} model introduced in~\cite{bogdanova2015detecting}.
We use the same AskUbuntu dataset to evaluate our models on a secondary dataset. 
There are two major differences between our approach and the works in \cite{bogdanova2015detecting} and \cite{rodrigues2017ways}.
First, we improve the performance of our model by computing the distance between title, body, and answers of the two \ku{}s, whereas \cite{bogdanova2015detecting} and \cite{rodrigues2017ways} only compute the similarity between title+body of the two \ku{}s. 
Second, the hybrid architectures developed by
\cite{rodrigues2017ways} is a complex CNN model along with 30k dense neural network followed by two hidden multi-layers.
However, our model uses shared layers bidirectional LSTMs with the limited number of parameters which results in a lightweight architecture. 

Question-Question similarity introduced in subtask B of SemEval-2017 Task 3
on Community Question Answering~\footnote{http://alt.qcri.org/semeval2017/task3/}~\cite{nakov2017semeval} is one of the closest topics to our task. 
Although this task contains multi-classes of relatedness between two questions (i.e., PerfectMatch, Related, Irrelevant), unlike our task, the problem is formulated as a re-ranking Question\_Question+Thread Similarity task. 
Various features were investigated to address Question-Question similarity introduced in subtask B of SemEval-2017 Task 3 such as neural embedding similarity features~\cite{goyal2017learningtoquestion} and Kernel-based features ~\cite{filice2017kelp}~\cite{galbraith2017talla}.
The winner of this task is~\cite{charlet2017simbow} which utilized soft-cosine similarity features within a Logistic Regression model. Note that we employ the similar soft-cosine features in our traditional SVM model. 

Duplicate detection between questions on \SO{} has been studied before.
\Comment{Zhang et al. proposed} An approach named \emph{DupPredictor} takes a new question as an input and tries to find potential duplicates of the question by considering multiple information sources (i.e., title, description and tags) \cite{zhang2015multi}.
\emph{DupPredictor} computes the latent topics of each question by using a topic model.
For each pair of questions, it computes four similarity scores by comparing their titles, descriptions, latent topics, and tags and then combined together to result in a new similarity score.
In another similar work, \cite{xu2016predicting} introduced a dataset for \ku{} relatedness and proposed a convolutional neural network for predicting the relatedness.
Unfortunately, the limited number of \ku{}s (KUs) were collected heuristically and tend to have low quality.
The presented dataset does not cover different parts of a \ku{}, instead, it merges title+body into a single sequence. Clearly, mixing all parts together does not provide an opportunity to perform an experiment on separate parts of \kua{}s independently.
Moreover, this dataset contains some extra information (signals) which leads to a biased dataset. As explained in ``Data Quality" section, we remove these unwanted clues from the data.

\section{Description of The Dataset}
Questions in the real world are supposed to have more relationships than only duplicate or non-duplicate. For example, one question in Stack Overflow talks about \emph{The time complexity of array function}\footnote{\url{https://goo.gl/dJwmuE}}, while another question is about \emph{How to find time complexity of an algorithm}\footnote{\url{https://goo.gl/S81BjE}}. These two questions are linked by Stack Overflow users as related but not duplicate. 

\subsection{Relatedness Between Knowledge Units} 
\label{sec:class_definition}
\begin{figure*}[h]
\centering
  \includegraphics[width=0.8\textwidth]{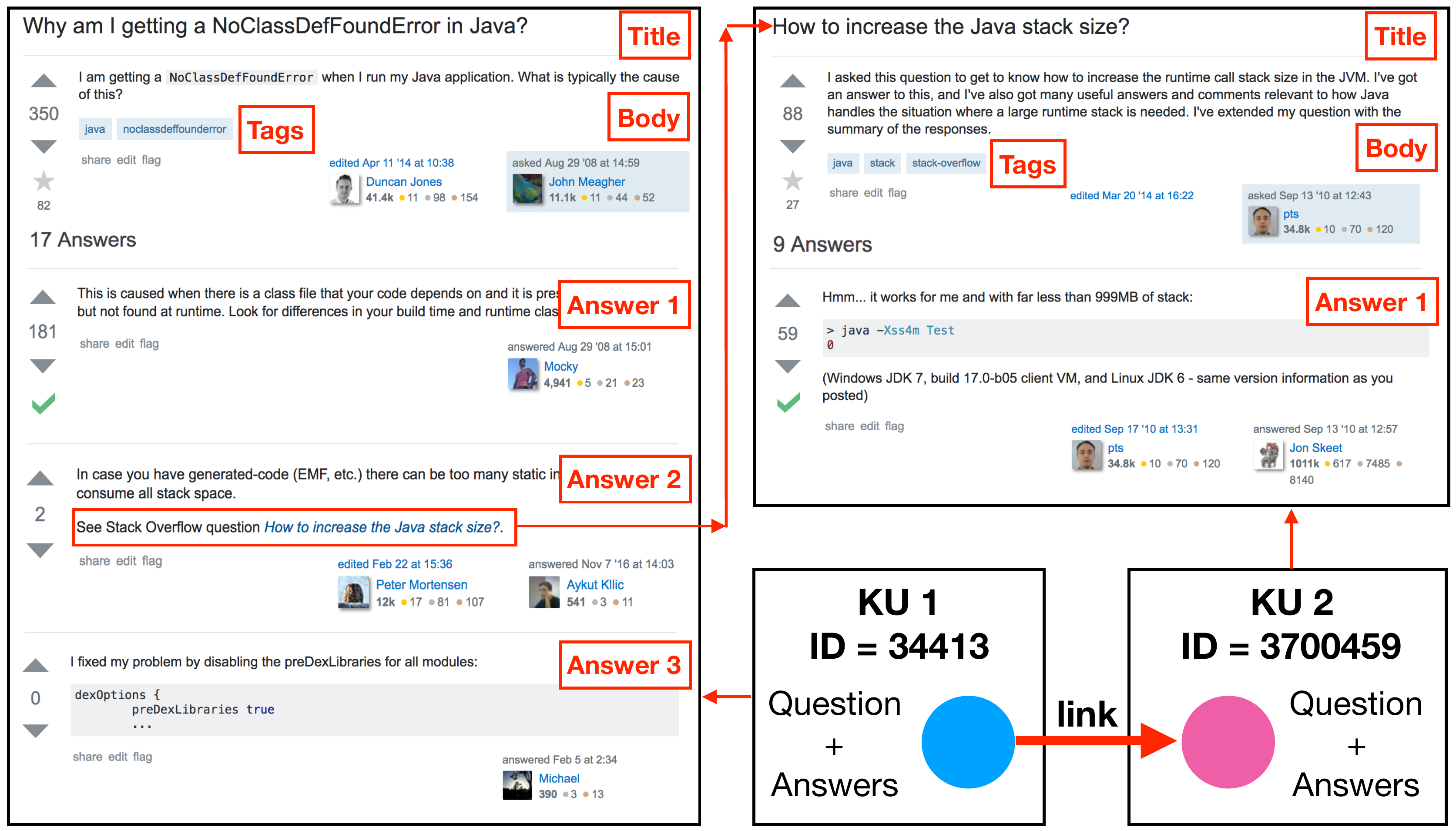}
  \caption{\small A pair of linkable knowledge units on Stack Overflow}
  \label{fig:example}
\end{figure*}

\noindent \Ku{}s often contain semantically-related knowledge, and thus they are linkable for different purposes, such as explaining certain concepts, approaches, background knowledge or describing a sub-step for solving a complex problem~\cite{Ye2016}. 
Figure~\ref{fig:example} shows an example of how \ku{}s are linked to each other on \SO{}.
One of the answers of a knowledge unit (short for \emph{KU1}) guides the asker to refer to another knowledge unit (short for \emph{KU2}) which is helpful to solve the problem.
These two \ku{}s are linked through URL sharing. 
URL sharing is strongly encouraged by \SO{} to link related \ku{}s~\cite{so1}.
A network of linkable knowledge units constitutes a \emph{\ku{} network} (KUNet) over time through URL sharing~\cite{Ye2016}.
Relationships between any two knowledge units in KUNet can be divided into four classes: \emph{duplicate}, \emph{direct}, \emph{indirect} and \emph{isolated} \cite{xu2016predicting}.
Duplicate KUs discuss the same question and can be answered by the same answer. Direct relatedness between KUs means that the content of one KU can help solve the problem in the other KU, for example, by explaining certain concepts, providing examples, or covering a sub-step for solving a complex problem.
Indirect relatedness means that contents of KUs are related but they are not immediately applicable to each other.
Isolated KUs are not semantically related.   
The order of relatedness of each class is \emph{duplicate} $>$ \emph{direct} $>$ \emph{indirect} $>$ \emph{isolated}.

\subsection{Dataset Creation} 
\label{sec:datasetbuild}
\begin{figure}[htbp]
\centering
  \includegraphics[width=0.5\textwidth]{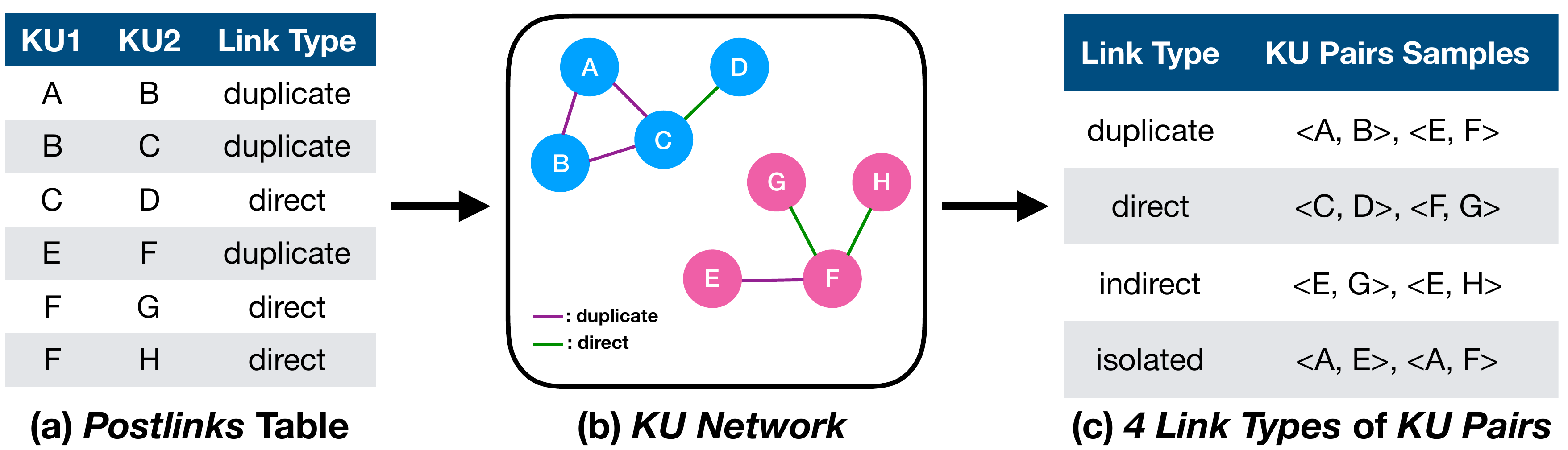}
  \caption{\small Overview of the data collection process}
  \label{fig:dataset}
\end{figure}

Figure~\ref{fig:dataset} depicts the steps that we took to create a relatedness dataset. We describe each step below.

\Part{Extract preliminary data from \SO{} data dump.}
We mainly focus on Java-related \ku{}s on Stack Overflow because
Java is one of the top-3 most popular tags in Stack Overflow~\footnote{https://stackoverflow. com/tags}.
Moreover, questions with this tag not only are about Java programming language, but they cover a broad spectrum of topics that Java technology provides, such as web and mobile programming, and embedded systems.
First, we extracted \emph{all} \ku{}s tagged by ``Java'' from \SO{} data dump.
Next, all \emph{duplicate} and \emph{direct} links between \ku{} pairs are extracted from \SO{} data dump.

\Part{Knowledge unit network.}
Knowledge unit network (KUNet)
is a network in which each KU is represented as a node and an edge between two nodes exists if a \emph{duplicate} or \emph{direct} link exists between the two corresponding KUs.
We construct a KUNet based on the extracted links from a table named \emph{PostLinks} from \SO{} data dump.

\Part{Identifying duplicate and direct pairs}
As shown in Figure~\ref{fig:dataset}(a), the link between ($A$ and $B$) and ($B$ and $C$) are labeled as a {\em duplicate}. 
We also consider a \emph{duplicate} link between $A$ and $C$ by transitivity. 
We apply transitivity rule until no new duplicate relation is found among \ku{}s. 

\Part{Identifying indirect and isolated pairs }
Four types linkable KU pairs are extracted from the KUNet based on their definitions.
{\em Indirect} KU pairs are pairs of nodes that are indirectly connected in the network.
More specifically, they are connected in the KUNet with a certain range of distance (in this case, length of shortest path $\in$ [2,5]), but the relationship between them belongs neither to duplicate nor direct. 
Finally, {\em isolated} KU pairs are pairs of nodes that are completely disconnected in the network. 


\subsection{Statistical Characteristics of the Dataset}	
Using the steps described in the previous section we created a dataset.
Table~\ref{tab:statistics} depicts the statistical characteristics of the dataset.
The dataset contains 160,161 distinct \ku{}s and 347,372 pairs of \ku{}s with four types of relationships.
Among all \ku{}s, $117,139$ (i.e., $73\%$) of them have at least one code snippet in their body. 
The average number of words in code snippets in body is $118.46$.
There are $318,491$ answers in our dataset and each \ku{} has $1.99$ answers on average.
 $140,122$ (i.e., $87\%$) of \ku{}s contain at least one answer and $90,672$ (i.e., $57\%$) of them contain one accepted answer.
Moreover, $96,707$ ($60\%$) of \ku{}s have at least one code snippet in their answers which means that more than half of solutions are code related.

\Part{Training, Development, and Test Sets}
We split the dataset into three parts, train, development, and test, to facilitate the development, and evaluation of classification models.
We assigned $60\%$ of \ku{}s to train set, $10\%$ to development set, and $30\%$ to test set.
To have the same number of KU pairs for each class,
by using under-sampling techniques, we make this dataset balanced.

\begin{table}[htbp]
\small
\centering
\caption{\small Brief statistics of the dataset}
\label{tab:statistics}
\resizebox{\columnwidth}{!}{%
\begin{tabular}{lll}
\hline
Scope                     & Indicator                                                     & Size          \\ \hline
                  & \# of distinct KUs                                            & 160,161       \\
\multirow{-2}{*}{Whole KU}                  & \# of four types of KU pairs                                            & 347,372       \\ \hline
Title                     & avg. \# of words in title                                          & 8.52          \\ \hline
                          & avg. \# of words in body(exclude code snippets)                    & 97.02         \\
                          & \# of distinct KUs whose body has at least one code snippet   & 117,139(73\%) \\
                          & avg. \# of code snippets in one body                         & 1.46          \\
\multirow{-4}{*}{Body}    & avg. \# of words in single code snippet in one body                   & 118.46        \\ \hline
                          & \# of distinct answers                                        & 318,491       \\
                          & avg. \# of answers within single KU                                  & 1.99          \\
                          & \# of distinct KUs contain at least one answer                & 140,122(87\%) \\
                          & \# of distinct KUs contain an accepted answer                 & 90,672(57\%) \\
                          & \# of distinct KUs whose answers has at least one code snippet & 96,707(60\%) \\
                          & avg. \# of words in an answer (exclude code snippets)              & 68.39         \\
                          & avg. \# of code snippets within one answer                        & 0.60          \\
\multirow{-8}{*}{Answers} & avg. \# of words in single code snippet                    & 81.98         \\ \hline
\end{tabular}%
}
\end{table}

\Comment{Amin: it's tricky
    \noindent\textbf{Tags }
    \Comment{
    Common tags between questions can be a good measure to asses the level of similarity \Fix{Bowen: We say tags can be a good indicator but we didn't use it in our experiment. It sounds a bit odd, do we need explain it?}.} Figure~\ref{fig:common_tags} shows the average number of common tags between pairs of questions for each relatedness class. It shows, questions in duplicate, direct and indirect classes share similar number of tags more than isolated classes. Thus our intuition is that differentiating between isolated pairs is relatively easier than other classes.
    Figure~\ref{fig:wordcloud} shows 40 most frequent tags in the whole datasets. Tags like Java, hibernate, android, maven are the most frequent through the entire dataset.
    
    \begin{figure}[htbp]
        \centering
        \begin{minipage}{0.45\textwidth}
            \centering
            \includegraphics[width=1.0\textwidth]{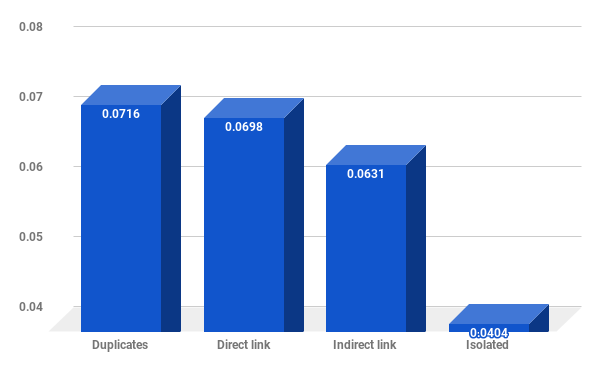}
            \vspace{-1cm}
            \caption{Average number of common tags between question pairs for each class}
            \label{fig:common_tags}
        \end{minipage}\hfill
        \begin{minipage}{0.45\textwidth}
            \centering
            \includegraphics[width=1.0\textwidth]{figs/wordcloud.png}
            \vspace{-1cm}
            \caption{wordcloud: most frequent tags in the entire dataset}
            \label{fig:wordcloud}
        \end{minipage}
    \end{figure}
}

\subsection{Instructions to Use The Dataset}
Table~\ref{tab:instruction} presents the overall structure of our dataset.
There are 24 attributes in our dataset for each pair of \ku{}s.
The first 23rd attributes include all the content of the first and second \ku{}s, they are id, title, body, accepted answer, answers, and tags.
The last attribute (i.e.,  Attr. Id =24) represents the relationship between the two \ku{}s (i.e., $<KU1, KU2, Relationship>$). More information is available at \url{https://anonymousaaai2019.github.io}


\begin{table}[htbp]
\small
\centering
\caption{\small The structure of the dataset}
\label{tab:instruction}
\resizebox{\columnwidth}{!}{%
\begin{tabular}{lll}
\hline
Attr. Id & Attr. Name        & Attr. Description                         \\ \hline
1        & Id                & KU Pair ($<KU1, KU2>$) Id                         \\  \hline
2/13        & q1/2\_Id            & Id of \emph{KU}'s Question on SO            \\            
3/14        & q1/2\_Title         & \emph{KU}'s Title                                       \\
4/15        & q1/2\_Body          & The text of \emph{KU}'s Body (Exclude Code Snippets)    \\
5/16        & q1/2\_BodyCode      & Code Snippets in \emph{KU}'s Body                    \\   
6/17        & q1/2\_AcceptedAnswerId      & Ids of \emph{KU}'s Accepted Answers on SO                    \\ 
7/18        & q1/2\_AcceptedAnswerBody      & The text of \emph{KU}'s Accepted Answer (Exclude Code Snippets)                    \\ 
8/19        & q1/2\_AcceptedAnswerCode      & Code Snippets in \emph{KU}'s Accepted Answer                    \\ 
9/20        & q1/2\_AnswersIdList & Ids of \emph{KU}'s Answers on SO                        \\
10/21       & q1/2\_AnswersBody   & The text of \emph{KU}'s Answers (Exclude Code Snippets) \\
11/22       & q1/2\_AnswersCode   & Code Snippets in \emph{KU}'s Answers         \\    
12/23       & q1/2\_Tags   & Tags of \emph{KU}         \\  \hline
24       & Class             & Relationship (i.e., \emph{duplicate}, \emph{direct}, \emph{indirect} or \emph{isolated})
\\ \hline
\end{tabular}%
}
\end{table}

\Comment{
\SO{} is a major source of knowledge for developers and system administrators.
A \ku{} (KU) in the \SO{} consists of a title, a question, tags chosen by the asker, and series of answers~\cite{xu2016predicting}. 
Users can use code snippet in any programming language to illustrate concepts, problems, or  solutions. 
They can use use links to other posts or links for further information.

\Fix{Sudipta:We already said almost the same thing in the introduction. Repetitive.}
}
\Comment{
\Fix{Reza: Doe we need to say the next sentence here?}
\SO{} posts are usually among the top results of search engines to the queries relevant programming and software operations.
}
\Comment{Xu et al. define a question together with its entire set of answers on \SO{} (short for SO) as a \emph{\ku} (short for KU)~\cite{xu2016predicting}.
}

\Comment{
As mentioned in Section~\ref{sec:datasetbuild}, we obtained corresponding question and answers for each \ku{} from the \emph{posts} table of \SO{} data dump based on the definition.
Our dataset contains 347,372 \ku{} pairs and each pair includes the information of two specific \ku{}s and the relationship between them, i.e., $<KU1, KU2, Relationship>$.
Because we focus on measuring semantic relatedness between two given \ku{}s, thus only semantic-related attributes are selected, i.e., title, body, answers.
We further extract text and code snippets from body and answers, separately. \FFix{Reza: it is part of "instruction to use", why we are talking about first we get .. then we get ... . let's just talk about different sections of the dataset. I think we need to bring the figure of different columns back, It shows how the DS is easy to use!}
Thus, for each \ku{}, we use two different kinds of attributes in this task, they are 1) text in title, body and answers, 2) code snippets in question's body and answers.
Thus, a total of five attributes are used in our approach.
To let researchers further explore it, we provide as more information as possible that there are 11 attributes for each \ku{} are provided in our dataset.
Please refer to the homepage of our dataset for more details\footnote{Dataset, \url{link here}}.
}

\Comment{
    Table~\ref{tab:SOexamples} shows real examples of \emph{duplicate}, \emph{direct}, \emph{indirect} KUs for a KU (to avoid confusion, we named it as ``original KU'') on \SO.
    Original KU (Id=4216745) is mainly about a question on \emph{\texttt{string} to {\tt date} conversion}.
    First, a KU (Id=6510724) has been recognized as a duplicate to the original KU by the Stack Overflow users;
    that is, they are actually the same question but formulated in different ways and they have the same answer.
    Second, a \emph{direct} KU (Id=4772425) is mainly about another question on \emph{change date format in a string}.
    One of the answers of the \emph{direct} KU\footnote{One answer of the \emph{direct} KU, \url{https://stackoverflow.com/a/4772461}} suggests asker to take a look on the original KU which is not the totally same question but it is strongly relevant and helpful to solve the original problem.
    Third, an \emph{indirect} KU (Id=47165086) which is directly link to one of the \emph{direct} linked KUs (i.e., KU Id=4772425) of the original KU.
    The \emph{indirect} KU is mainly about \emph{convert datetime in jmeter using beanshell sampler} which may not directly help for solving the problem of original KU.
    However, the \emph{indirect} KU has relevant knowledge with the original KU which is helpful for developers to learn the breadth of knowledge, especially for junior developers~\cite{xu2016predicting}.
    
    \noindent
\begin{table*}[h]
\small
\centering
\vspace{-0.2cm}
\caption{\small Examples of \emph{duplicate}, \emph{direct} and \emph{indirect} Knowledge Units from Stack Overflow}
\label{tab:SOexamples}
\footnotesize
\begin{tabular}{lp{8cm}l}
\hline
     \emph{Original} KU (\url{https://stackoverflow.com/questions/4216745/})  \\ \hline
\textbf{Title:} Java string to date conversion   \\
\begin{tabular}[c]{@{}l@{}} \textbf{Body:} Can somebody recommend the best way to convert a string in the format 'January 2, 2010' to a date \\ 
in java? Ultimately, I want to break out the month, the day, and the year as integers so that I can use: \\
\begin{lstlisting}[language=java] 
Date date = new (*\bfseries Date*)();
date.setMonth()..
date.setYear().. 
date.setDay()..
date.setlong currentTime = date.getTime();
\end{lstlisting} \\
to convert the date into time.
\end{tabular} \\ \hline

     \emph{Duplicate} KU (\url{https://stackoverflow.com/questions/6510724/})  \\ \hline
\textbf{Title:} how to convert java string to Date object      \\
\begin{tabular}[c]{@{}l@{}} \textbf{Body:} 
I have a string \\
\begin{lstlisting}[language=java] 
String startDate = "06/27/2007";
\end{lstlisting} \\
now i have to get Date object. My DateObject should be the same value as of startDate.\\
I am doing like this \\
\begin{lstlisting}[language=java] 
DateFormat df = new SimpleDateFormat("mm/dd/yyyy");
(*\bfseries Date*) startDate = df.parse(startDate);
\end{lstlisting} \\
But the output is in format \\
\begin{lstlisting}[language=java] 
Jan 27 00:06:00 PST 2007.
\end{lstlisting}
\end{tabular} \\ \hline

     \emph{Direct} KU (\url{https://stackoverflow.com/questions/4772425/})  \\ \hline
\textbf{Title:} Change date format in a Java string     \\
\begin{tabular}[c]{@{}l@{}} \textbf{Body:} 
I've a String representing a date. \\
\begin{lstlisting}[language=java] 
String date_s = "2011-01-18 00:00:00.0";
\end{lstlisting}\\
I'd like to convert it to a Date and output it in YYYY-MM-DD format.
\begin{lstlisting}[language=java] 
2011-01-18
\end{lstlisting} \\
How can I achieve this? \\
Okay, based on the answers I retrieved below, here's something I've tried: \\
\begin{lstlisting}[language=java] 
String date_s = " 2011-01-18 00:00:00.0"; 
SimpleDateFormat dt = new SimpleDateFormat("yyyyy-mm-dd hh:mm:ss"); 
(*\bfseries Date*) date = dt.parse(date_s); 
SimpleDateFormat dt1 = new SimpleDateFormat("yyyyy-mm-dd");
System.out.println(dt1.format(date));
\end{lstlisting} \\
But it outputs 02011-00-1 instead of the desired 2011-01-18. What am I doing wrong?
\end{tabular} \\ \hline

     \emph{Indirect} KU (\url{https://stackoverflow.com/questions/47165086/})  \\ \hline
\textbf{Title:} How to convert datetime in jmeter using beanshell sampler    \\
\begin{tabular}[c]{@{}l@{}} \textbf{Body:} 
I have timestamp for one of my http sampler in following format \\
\begin{lstlisting}[language=java] 
Tue Nov 07 10:28:10 PST 2017
\end{lstlisting} \\
and i need to convert it to in following format \\
\begin{lstlisting}[language=java] 
11/07/2017 10:28:10
\end{lstlisting} \\
i tried different approaches but don't know what am i doing wrong.Can anyone help me on that.Thanks.
\end{tabular} \\ \hline
\end{tabular}
\end{table*}

}

\section{Quality Control}\label{sec:quality}
\subsection{Data Cleaning}
We perform three operations to further improve the quality of our dataset.
 Natural language and programming language snippets are mixed in the text. To deal with this, first, we extract programming language snippets (aka. code snippets) from HTML formatted text by using the regular expression $\langle pre \rangle  \langle  code \rangle(.*?)\langle/code\rangle \langle/pre\rangle$
Note that, it is possible that multiple code snippets exist in body or multiple answers of one \ku{}, so we store them into a list.
Next, since text attributes (e.g., body, answer body) provided by Stack Overflow data dump are in HTML format, we clean the content by removing HTML tags and escape characters, e.g., $\langle p \rangle \langle /p \rangle$, $\&\#xA;$ and $\&lt;$.
Second, we observe and remove some extra information added by Stack Exchange API that can be considered as a signal. For example, at the beginning of the body content of some duplicate and direct questions, it includes the string \texttt{Possible Duplicate:}, followed by the topic content of the possible duplicate question.
The inclusion of signals in training can result in a biased dataset and unreliable models. This problem was first observed by~\cite{silva2020} in AskUbuntu dataset.

Third, we found that there is an overlap between some \emph{duplicate} and \emph{direct} links in the Stack Overflow data dump, since it provides \ku{} pairs as long as two \ku{}s are linked through URL sharing.
To solve this, if a link belongs to \emph{duplicate} and \emph{direct} at the same time, we label it as a  \emph{duplicate}.

\subsection{User Study}
This dataset is extracted from Stack Overflow forum that is managed and maintained by volunteer domain experts who serve as moderators and contributors. 
Links between knowledge units (i.e., Stack Overflow posts) are validated in a crowdsourced process by domain experts.
To asses the reliability of the crowdsourced process and our data collection procedure,  
we perform a user study.
We ask three experts (who are not authors of this paper) to label relationships between pairs of knowledge units that we have in our dataset. 
The participants analyze a statistically significant sample size (i.e., 96 pairs) that is representative of the population of knowledge units in our dataset (at 95\% confidence level, and 10\% margin or error). 
Each participant can provide his/her assessment of the degree of relatedness of two knowledge units in a 4 point Likert scale: 1 (unrelated/isolated), 2 (indirect), 3 (direct), and 4 (duplicate). The user study highlights that the participant labels are the same as the labels in our dataset 82\% of the time. The average absolute difference between the Likert scores and the labels in our dataset is only 0.2 (out of 4). This highlights that the links in our dataset are of high-quality.
\section{Method}
In this section, we describe models to predict relatedness between \ku{}s.
We extensively explore different neural network and traditional models for this task and report the best-performing models. 
First, we investigate a BiLSTM architecture which progressively learns and compares the semantic representation of different parts of two \ku{}s.
The description of our model is presented in the next section.
We then compare the BiLSTM model with a support vector machine model.  
We also apply these models to a closely similar task, duplicate detection in AskUbuntu, and compare the results with the state-of-the-art models in that task.

\subsection{Data Pre-processing}
We apply some simple pre-processing steps on all text parts, Title, Body and Answers.  
Since there are many technical terms in \SO, we apply more specific pre-processing steps:  
First, we split words with punctuation marks.
For example, \texttt{javax.persistence.Query javax\_query} changes to \texttt{javax persistence Query javax query}.
Then, we split camel case words, for example, \texttt{EntityManage} is changed to \texttt{Entity Manage}.
In the end, we take several standard steps in preprocessing data including: normalizing URLs and numbers, removing punctuation marks and stop-words, and changing all words to lowercase.

\section{LSTM Model}\label{sec:deepModel}
\label{section:DL}

We use bidirectional long short-term memory (BiLSTM) ~\cite{hochreiter1997long} as a sentence encoder to capture long-term dependencies in forward and backward directions. In a simple form, an LSTM unit contains a memory cell with self-connections, as well as three multiplicative
gates to control information flow. 
Given input vector $x_t$, previous hidden outputs $h_{t- 1}$, and previous cell state $c_{t- 1}$, LSTM units operate as Figure~\ref{fig:formaula},
where $i_t$, $f_t$, $o_t$ are input, forget, and output gates, respectively. 
The sigmoid function $\sigma()$ is a soft gate function controlling the amount of information flow. 
$W_s$ and $bs$ are model parameters to learn.

\begin{figure}

\begin{gather*}\label{formula_LSTM}
    X= \left[ \begin{array}{l}
    x_t\\ h_{t -1} \end{array}\right] \\
    i_t = \sigma(W_{iX}X + W_{ic}ct - 1 + b_i)
    \\
f_t = \sigma(W_{fX}X + W_{fc}ct - 1 + b_f) \\
o_t = \sigma(W_{oX}X + W_{oc}c_t - 1 + b_o) \\
c_t = f_t \odot c_{t-1} + i_t \odot tanh(W_{cX}X + b_c) \\
h_t = o_t\odot tanh(c_t) \\
\end{gather*}
\vspace{-12mm}
    \caption{LSTM  Unit}
    \label{fig:formaula}

\end{figure}

\begin{figure}[h!]
\centering
  \includegraphics[width=0.45\textwidth]{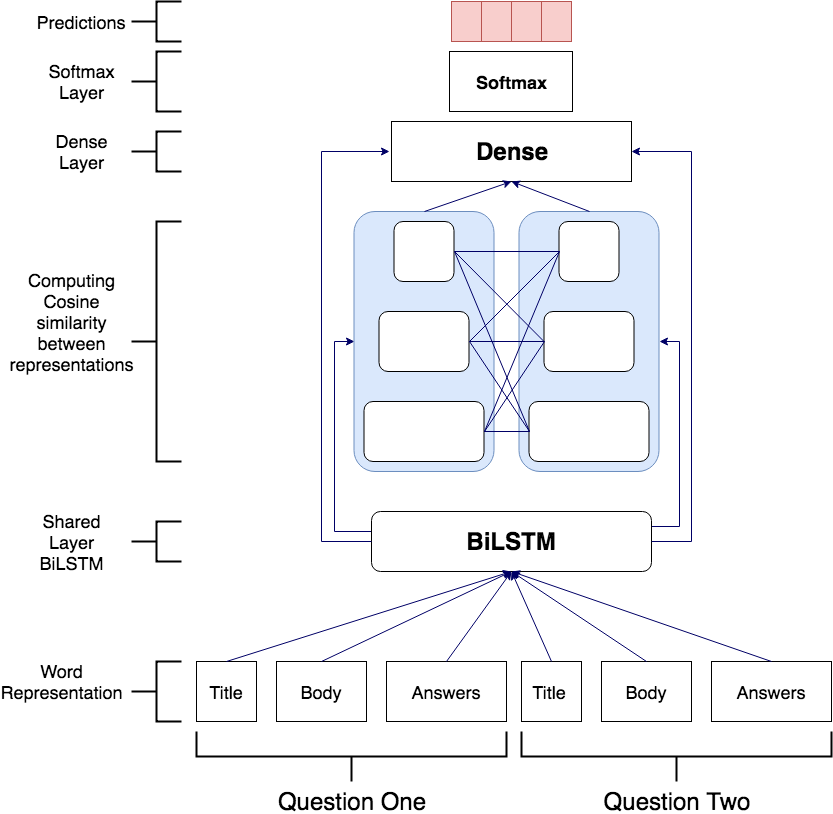}
  \caption{Main architecture of \dotbilstm{}}
  \label{fig:main}
\end{figure}

Figure~\ref{fig:main} describes the overall architecture of the BiLSTM model (\dotbilstm{}). 
Unlike previous studies
(i.e. \cite{rodrigues2017ways}\cite{bogdanova2015detecting}), 
this model utilizes the information in Title, Body and Answers parts of each \ku{}. 
Each word ($w_i$) is represented as a vector, $\mathbf{w} \in \mathbb{R}^d$ , looked up into an embedding matrix, $\mathbf{E} \in \mathbb{R}^{d\times|V|}$. 
A shared layer BiLSTM as a sentence encoder takes all the six inputs, embeds and transforms them into fixed-sized vectors. 
Then in order to compute the distance between each two \ku{}s, we compute the inner dot product between all the three representations of the first \ku{} and all three representations of the second \ku{}. As a result, it maps a pair of \ku{}s into a low dimensional space, where their distance is small if they are similar. 
In the next step, we concatenate computed values together. 
Our results show that concatenating the BiLSTM representations at the last layer increases the performance slightly.
We feed these values to a fully-connected layer followed by a ReLU activation function, a dropout layer and then a \emph{SoftMax} output layer for classification.
The objective function is the Categorical cross-entropy objective over four class target labels.

\subsection{Implementation Details (\dotbilstm{})}
This section describes implementation details which are empirically chosen after running several models with different values and keeping the one that gives us the best results in the validation set.  

We initialize word embeddings with pre-trained GloVe~\cite{pennington2014glove} vectors of size 300. 
Compared to pre-trained Google news word2vec~\cite{mikolov2013efficient} and word embedding trained on \SO{}, GloVe performed slightly better in this task.
We choose the size of each sentence based on the average size over the training set. 
Titles are truncated or padded to 10 words, bodies to 60 words and answers to 180. 
BiLSTMs with 128 units is used as the encoder. 
In our experiments, we observed that using shared parameters for BiLSTMs boosts the model. 
The network uses Adam optimizer~\cite{kingma2014adam}, and the learning rate is set to $0.001$. 
The last layer is a dense layer with ReLu activation and 50 units.
In order to have a better training and force the network to find different activation paths which leads to a better generalizing, 
a dropout layer with the rate of 0.2 is used.
All the models are trained for 25 epochs and the reported test accuracy corresponds to the best accuracy obtained on the validation set. 

\section{SVM model}
\label{sec:baseline}

In this section, we explain the design of \SSVM{}, an SVM model for question relatedness task. 
We investigate different features as well as different data selections to achieve the best possible results.

\label{section:features}
We extract three types of features from \ku{}s: 
\emph{Number of common $n$-grams} which is simply the number of common word $n$-grams, and common character $n$-grams in a pair of text sequences. 
\emph{Cosine similarity measure} to determine the similarity between two vectors~\cite{kenter2015short,levy2015improving}.
This feature is obtained by TF-IDF weighting, computed over the training and development datasets.
And, \emph{Soft-cosine similarity measures} that unlike the traditional cosine similarity, takes into account word-level
relations by computing a relation matrix~\cite{sidorov2014soft}. 
Given two N-dimension vectors a and b, the soft cosine similarity is calculated 
as follows.
\begin{equation}\label{softcosine}
soft-cosine(a,b) = \frac{\sum_i,_j^Na_i m_{ij} b_j}{\sqrt{\sum_i,_j^N a_i m_{ij} a_j}\sqrt{\sum_i,_j^N b_i m_{ij} b_j}}
\end{equation}
Unlike cosine similarity, soft-cosine similarity between two texts without any words in common is not null as soon as the two texts share related
words. For computing the matrix $M$, we followed the same implementation presented in ~\cite{charlet2017simbow}, the winner of SemEval-2017 Task 3, Question-Question similarity. We create three variants of soft-cosine similarity feature. 
One is computed based on Levenshtein distance  (\softLev{}),
and the other two features are based on two different word embeddings: Google News pre-trained word2vec~\cite{mikolov2013efficient}(\softgoog{}) and \SO{} domain-specific word2vec (\softSO{}).

\subsection{Implementation Details (\SSVM)}

We build an SVM model with the linear kernel using \texttt{sklearn} package~\cite{pedregosa2011scikit}.
In total, for each \kua{} pair, we extract ten different hand-crafted features: three common word $n$-grams (for $n$=1,2 and 3), three common character $n$-grams (for $n$=3,4 and 5), cosine similarity and three soft-cosine similarity features \softSO{}, \softLev{} and \softgoog{}.
We compute the features between titles, bodies and answers separately.
For computing \softSO{}, we train word2vec on text parts of the dataset using skip-gram model~\cite{mikolov2013distributed}
with vectors dimension 200 and minimum word frequency of 20.

\subsection{Feature Selection}
\begin{figure}[htbp]
\centering
\begin{minipage}{0.45\textwidth}
    \centering
    \includegraphics[width=1\textwidth]{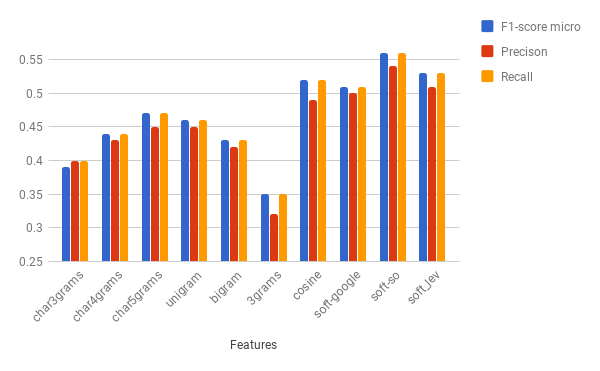}
    \vspace{-5mm}
    \caption{\small Performance of SVM models using individual features}
\label{fig:featureText}
\end{minipage}\hfill
\end{figure}

In this section, we compare and select important features by building SVM models using each feature separately. 
As shown in Figure~\ref{fig:featureText}, cosine and three Soft-cosine features outperform other features. 
Therefore, we choose cosine similarity, \softSO{}, \softgoog{}, and \softLev{}, as the final feature set in the \SSVM{} because they perform better than other features.  

To compare and select the important text selection parts, we build the SVM model by considering only title, body or answers. 
As shown in Table~\ref{table:textSelection}, 
the model with different parts perform similarly and the 
best performance is achieved when we consider all three, title, body and answers. 
\begin{table}[htbp]
\small
\centering
\caption{\small 
Results of choosing different text selections.
}
\label{table:textSelection}
 \resizebox{\columnwidth}{!}{
\begin{tabular}{|l|l|l|l|l|}
\hline
	Text selection/metrics      & F-micro & Precision & Recall \\ \hline
	Title                                &       0.47         &     0.44      &   0.48     \\ \hline
	Body                               &     0.51           &    0.49       &   0.51     \\ \hline
	Answers                            &      0.51          &   0.5        &      0.51  \\ \hline
	Title, Body, Answers                       &      0.59          &     0.58      &   0.59     \\ \hline
\end{tabular}
}
\end{table}

\section{Results and Discussion}
\Part{Analysis of Results  }
Table~\ref{tab_results} compares results for both \SSVM{} and \dotbilstm{} on \SO{} dataset.
Comparing the obtained results, we realize that \dotbilstm{} substantially outperforms \SSVM{} by more than 16 absolute percentage point in F-micro.
This suggests that the BiLSTM model can utilize the large amount of training data in \SO{} dataset and predict the relatedness between \ku{}s more effectively than our traditional model. 

\begin{table}[]
\centering
\caption{\small Results for \SSVM{} and \dotbilstm{} models}
\label{tab_results}
\begin{tabular}{|l|l|l|l|}
\hline
Model/Metrics  & F-micro & Precision & Recall \\ \hline
\SSVM{}       & 0.59           & 0.58      & 0.59   \\ \hline
\dotbilstm{}    &  0.75           & 0.75       &0.75   \\ \hline
\end{tabular}
\end{table}

Tables~\ref{table:classes} shows F-micro scores for predicting individual classes.
Comparing results of the individual classes, \dotbilstm{} performs better than \SSVM{} in predicting Isolated, Duplicate and Indirect classes. 

\begin{table}[htbp]
\centering 
\caption{\small Comparing the results (f-score) of \SSVM{} and \dotbilstm{} models}
\label{table:classes}
\resizebox{\columnwidth}{!}{%
\begin{tabular}{|l|l|l|l|l|l|}
\hline
Models/Classes     & Duplicate & Direct &       Indirect &         Isolated & Overall: Micro \\ \hline
\SSVM{}    & 0.53      & 0.57        & 0.44          & 0.79     & 0.59           \\ \hline
\dotbilstm{} &   0.92         & 0.55        & 0.67          & 0.87     & 0.75           \\ \hline
\end{tabular}
}
\end{table}


\Part{Reformulating the problem to the binary format of \emph{Duplicate Detection}:  }
For having a better comparison between our task and other typical duplicate/non-duplicate classification studies (some mentioned in ``Related Work"), we \emph{reformulate} the task to Duplicate Question Detection (DQD) and report the results of our models in the 2-class scenario.
DQD is to predict if two given \ku{}s are either duplicate or non-duplicate.
To evaluate the models under the DQD scenario, we need to map four relatedness classes into two Duplicate and Not-duplicate classes.
We consider duplicate class from the original dataset as \emph{duplicate} and the rest as \emph{non-duplicates} instances.
To address the imbalanced class problem,
we apply under-sampling techniques for non-duplicate class. More precisely, we randomly choose instances 
from all other three classes (direct, indirect and isolated) to have an equal number of both classes.
By reformulating the task from multi-class to binary classification task, we expect our models to achieve higher results. 
We evaluate both \dotbilstm{} and \SSVM{} models using the reformulated dataset.
\dotbilstm{} and \SSVM{} prediction performances increase to 0.91 and 0.70 f-score respectively. As we expected, by having two classes instead of four, in a relatively simpler problem, \dotbilstm{} and \SSVM{} results increase by 16\% percent and 11\% respectively. 


\Part{Comparing with AskUbuntu Dataset:  }\label{sec:askubuntu}
We take a further step and expand our work by investigating AskUbutu DQD dataset for two specific reasons: (1) to show the robustness of the used models, and  
(2) To show the challenging nature of the proposed dataset on \SO{} compared to others.
We expect to observe a different behavior of our models on this data due to the different nature and structure. 
For example, unlike \SO{}, the inputs of AskUbuntu dataset are only limited to title+body of each question. 
Moreover, AskUbuntu data contains a fewer number of instances, that is 24K pairs for training, 6K for testing and 1K for validation part.
We use the cleaned version of AskUbuntu dataset (without signal) prepared by~\cite{rodrigues2017ways}.
Using the same splitting used in~\cite{rodrigues2017ways}, our both models perform similarly. \dotbilstm{} model achieves 0.88 f-score and 0.87 accuracy, and \SSVM{} model achieves 0.90 f-score and 0.90 accuracy. 
Our models outperform the state-of-the-art \emph{Hybrid DCNN} model on this dataset introduced in~\cite{rodrigues2017ways} with the accuracy of 0.79.
This shows that not only our lightweight BiLSTM and traditional SVM model perform well in the \SO{} dataset, these models also outperform the complex \emph{Hybrid DCNN} model on AskUbuntu dataset. 
Note that in order to evaluate the models on this dataset, we need to customize models to only have 2 inputs (title+body pairs). 

\Part{More To Explore:  }
In our experiments, we purposely confined our models to only utilize information in \emph{Title}, \emph{Body} and \emph{Answers}. However, relying on other parts of the dataset like \emph{BestAnswer}, \emph{Tags} and \emph{Code parts} can boost performance further for this task. As future work, we intend to investigate to utilize code parts as they are considered as informative resources about the content of the \ku{}s.

\section{Conclusion}
This paper presents the task along with a large-scale dataset for identifying relatedness of \ku{} (question thread) pairs in \SO{}.
We reported all the steps for creating this dataset and a user study to evaluate the quality of the dataset. 
We devised two models, \dotbilstm{} and \SSVM{} for this task and their performances for future evaluations.
We also compared the performance of  \dotbilstm{} and \SSVM{} models  with the state-of-the-art model on AskUbuntu dataset and found that these models outperform the state-of-the-art model. 
We made the dataset and models available online.

\bibliography{bib}
\bibliographystyle{aaai}

\end{document}